\documentclass[12pt]{article}
\usepackage{times}
\usepackage{graphicx}
\usepackage{listings}
\usepackage{wrapfig}
\usepackage[usenames,dvipsnames]{xcolor}

\newtheorem{defin}{Definition} 

\newenvironment{sciabstract}{%
\begin{quote} \bf}
{\end{quote}}

\newcounter{lastnote}

\title{Implementing a Wall-In Building Placement in StarCraft with Declarative Programming}

\author
{Michal \v{C}ertick\'{y}\\
\\
\normalsize{Agent Technology Center, Czech Technical University in Prague}\\
\normalsize{\tt michal.certicky@agents.fel.cvut.cz}
}

\date{June 2013}

\begin{document} 
\hyphenation{ty-pi-cal se-ve-ral in-di-vi-dualrea-so-ning ene-my dec-la-ra-ti-ve du-ring rea-so-ning buil-dings}


\baselineskip24pt


\maketitle


\begin{sciabstract}
In real-time strategy games like StarCraft, skilled players often block the
entrance to their base with buildings to prevent the
opponent's units from getting inside. This technique, called ``walling-in", is a vital part of player's
skill set, allowing him to survive early aggression. However, current 
artificial players (bots) do not possess this skill, due to numerous 
inconveniences surfacing during its implementation in imperative languages 
like C++ or Java. 
In this text, written as a guide for bot programmers, we address the problem 
of finding an appropriate building placement that would block the entrance to player's base, 
and present a ready to use declarative solution employing the paradigm of answer set programming 
(ASP). We also encourage the readers to experiment with different declarative 
approaches to this problem.
\end{sciabstract}

{\bf Keywords:} StarCraft, Real-time Strategy, Answer Set Programming, Wall-In, BWAPI


\section{Introduction}

StarCraft\footnote{StarCraft and StarCraft: Brood War are trademarks of Blizzard Entertain ment, 
Inc. in the U.S. and/or other Countries.} is a popular computer game representing a Real-time Strategy (RTS) 
genre. In a typical RTS setting, players (either human or artificial) are in control of various structures (buildings) 
and units which they order to gather
resources, build additional units and structures, or attack the opponent.
RTS games are in general a very interesting domain for Artificial Intelligence (AI) 
research, since they represent well-defined complex adversarial systems and can be divided into many 
interesting sub-problems \cite{Buro2004}.

Expert knowledge about such a complex environment is extensive and hard-coding the reasoning over it in an 
imperative programming language like C++ or Java may in some cases prove time-consuming and inconvenient. 
Various declarative knowledge representation paradigms are well-suited for some of the subproblems of RTS AI, 
and their corresponding state-of-the-art solvers can often do most of the work for us. 

In this text, we address the subproblem of finding an appropriate building placement that would effectively block
the entrance to player's base region. Skilled human players often block the narrow entrance (chokepoint) to their
base with their own structures in order to prevent the enemy units from getting inside. This technique is a vital
component of any StarCraft player's skill set, allowing him to survive the early phase of the game against aggressive
opponents. 

Over past few years, we have seen a great amount of research conducted in the area of artificial intelligence for
RTS games, especially StarCraft, thanks to the introduction of the BWAPI framework \cite{bwapi}. Many of relevant publications deal with various machine learning approaches, either for 
micro-management in combat \cite{Wender-Watson-12, Rathe-12}
or for macro-economic or strategic tasks \cite{Dereszynski-11, Synnaeve-12, Certicky-13}.
Others solve the opponent modelling \cite{Fjell-12} or optimization problems over possible build orders \cite{Churchill-10}.
However, there has been no publications dealing with the problem of wall-in building placement so far. Even though there
is a large number
of high-quality artificial players (bots) competing in long-term tournaments like SSCAI\footnote{\tt http://sscaitournament.com/}, or 
shorter events on conferences like AIIDE or CIG, they seem to perform poorly against early aggression, since {\em none of them}
is able to use buildings to block the entrance to their base. We write this text in hope that it will serve as a guide for bot creators, 
allowing them to solve this problem quickly and effortlessly.

After describing the problem at hand more closely in section \ref{section:proglem-description}, we will briefly
outline the semantics of answer set programming (ASP), a paradigm of logic programming employed in our 
prototypical problem encoding described in detail in section \ref{section:encoding}.

\section{The Problem Description}\label{section:proglem-description}

The problem of wall-in building placement can be seen as a constraint satisfaction problem (CSP) \cite{Dechter-88}. 
Typically, a CSP is defined as a triple $\langle X, D, C \rangle$, where $X$ is a set of variables, $D$ a set of values to be assigned to 
them and $C$ is a set of constraints that need to be satisfied in any valid solution (assignment).
In our case, variables correspond to individual buildings that we want to use in our blockade (wall) and values are all 
the tile positions\footnote{The notion of a ``{\em tile position}" will be explained further in the text.}
around the chokepoint. In other words, we need to assign a placement to every building 
we are going to build, such that all of the following constraints are satisfied:
\begin{itemize}
	\item Every building {\em can be} built on its assigned location (this depends on the terrain).
	\item Buildings {\em do not overlap}.
	\item There will {\em not be} any {\em walkable path} leading from ``outside" region to our base, after the buildings are constructed.
\end{itemize}

\begin{figure}[h!]
  \centering
    \includegraphics[width=0.9\textwidth]{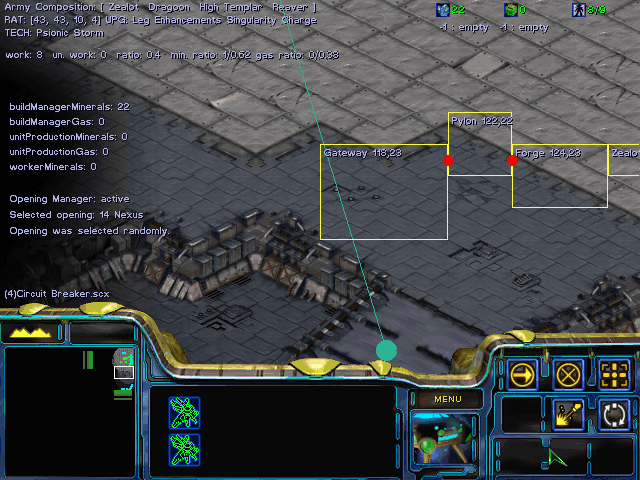}
  \caption{Example of wall-in placement as a Protoss. The wall consists of a Gateway, Forge and 
  Pylon structures and a Zealot unit. In CSP terms, variables from $X = \left\{ Gateway, Pylon, Forge, Zealot \right\}$
  are assigned the values of $(118,23), (122,22), (124,23)$ and $(126,23)$ respectively.}
  \label{fig:wall-toss1}
\end{figure}

{\bf Note (Protoss):} The situation is slightly more complicated when our bot plays as Protoss (one of three playable races 
in StarCraft). We need one more constraint about all the buildings being powered by a Pylon, and we do not want to wall-in
completely, because otherwise we would not be able to get out of the base (in contrast, the Terran bots can lift their 
buildings to get out and Zerg bots generally do not build walls at all). Therefore, in addition to buildings, we need to include one
unit in our wall (typically a Zealot). Fortunately, this is simply solved by adding one extra variable to $X$ (see figure 
\ref{fig:wall-toss1}).

{\bf Note (Tiles):} For the purposes of building placement, the map in StarCraft is divided 
into a grid of square $32 \times 32$px ``{\em build tiles}". The buildings can only be placed onto discrete positions
within this grid (unlike units, which can move around freely). Every building then occupies several tile positions, 
depending on its size, while its position is usually denoted by the coordinates of the {\em top left} tile occupied by it.

\begin{figure}[h!]
  \centering
    \includegraphics[width=0.45\textwidth]{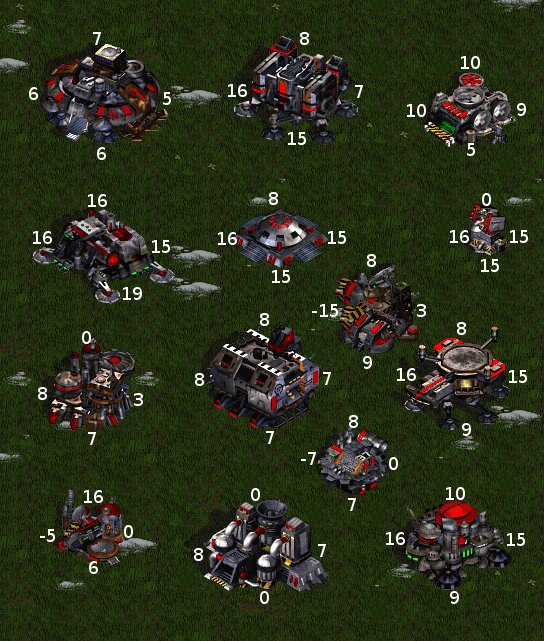}
    \includegraphics[width=0.45\textwidth]{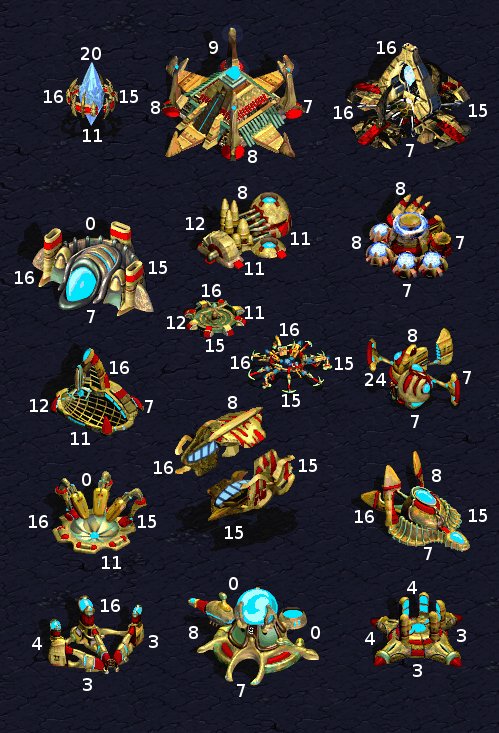}
  \caption{Complete enumeration of gap sizes around Terran and Protoss buildings. Negative gap sizes mean
  that a given building blocks an area outside its assigned tiles.}
  \label{fig:gaps}
\end{figure}

The game engine of StarCraft causes one more complication: the buildings do not block the entire area 
covered by the tiles they occupy. For example (see figure \ref{fig:wall-toss1}), a Forge building occupies an area of $3 \times 2$ 
tiles. However, there is a $12$px wide walkable gap at the
left side, and an $11$px wide gap at the right side of the Forge. The sizes of these gaps are hard-coded 
in the game engine and are different for every building type. See the gap sizes for all the Terran and Protoss 
building types in figure \ref{fig:gaps}. Some walls might contain gaps between the buildings that
are wide enough for smaller units (e.g. Zerglings with their $16 \times 16$px dimension) to walk through. 
For example, there is a $27$px wide gap between the Forge and the Pylon in figure \ref{fig:wall-toss1}.
If gaps like this cannot be avoided, they need to be blocked afterwards by additional units.

This brings us to the {\em optimization} part of our problem. A CSP defined like this often has 
more than one valid solution. Constraints may be satisfied by various assignments of tile positions
to buildings. We can, however, define and select the best possible valid assignment based on how wide the
gaps between individual buildings are.

Both the constraint satisfaction and optimization problems can be solved simultaneously by 
a certain paradigm of logic programming, called the {\em Answer Set Programming} or ASP (details in the following 
section). This, together with the existence of effective solvers, is the reason why we have chosen the ASP 
for our implementation. However, we emphasize that various 
other declarative approaches (different paradigms of logic programming, constraint programming, etc.) can
undoubtedly be used as well. The ideas and solutions described here should be easily transferable to other 
languages and tools.

To implement a wall-in placement generator in an imperative programming language, one would 
need to solve the following subproblems one by one:
\begin{itemize}
	\item Search over the space of all possible building placements.
	\item Implement a proper pathfinding algorithm, which is not so easy with different gaps between buildings.
	\item Call this algorithm for every valid placement to determine if the wall is indeed tight.
\end{itemize}
In contrast to that, when taking a declarative approach, we only need to define the problem correctly. 
The search over a solution space, constraint testing and optimization is taken care of by the solver (in our case,
an ASP solver {\em clingo} \cite{Gebser-08}).

\section{Answer Set Programming}\label{section:asp}

Answer Set Programming \cite{Gelfond-Lifschitz-1988, Gelfond-Lifschitz-1991, Baral-2003}
has lately become a popular declarative problem solving paradigm
with growing number of applications. 

The original language associated with ASP allows us to formalize various kinds of common sense
knowledge and reasoning, including constraint satisfaction and optimization. 
The language is a product of a research aimed at defining a formal semantics for
logic programs with default negation \cite{Gelfond-Lifschitz-1988}, and was extended to allow also 
a classical (or \emph{explicit}) negation in 1991 \cite{Gelfond-Lifschitz-1991}.
An ASP \emph{logic program} is a set of \emph{rules} of the following form:
\begin{displaymath}
h \leftarrow l_1, \dots, l_m,\ not\ l_{m+1}, \dots\ not\ l_n.
\end{displaymath}
where $h$ and $l_1, \dots, l_n$ are classical first-order-logic literals and $not$ denotes a default negation. Informally, such rule
means that ``if you believe $l_1, \dots, l_m$, and have no reason to believe any of $l_{m+1}, \dots, l_n$, then
you must believe $h$". The part to the right of the ``$\leftarrow$" symbol ($l_1, \dots, l_m,\ not\ l_{m+1}, \dots\ not\ l_n$) 
is called a \emph{body}, while the part to the left of it ($h$) is called a \emph{head} of the rule. 
Rules with an empty body are called \emph{facts}. Rules with empty head are called \emph{constraints}.

The ASP semantics is built on the concept of \emph{answer sets}. Consider a logic program $\Pi$ and a consistent 
set of classical literals $S$. We can then get a subprogram called \emph{program reduct of $\Pi$ w.r.t. set S} (denoted $\Pi^S$) by removing each 
rule that contains $not\ l$, such that $l \in S$, in its body, and by removing 
every ``$not\ l$" statement, such that $l \notin S$. We say that a set of classical literals $S$ is \emph{closed} under a 
rule $a$, if holds $body(a) \subseteq S \Rightarrow head(a) \in S$. 

\begin{defin}[Answer Set]
Let $\Pi$ be a logic program. Any minimal set $S$ of literals \emph{closed} 
under all the rules of $\Pi^S$ is called an \emph{answer set} of $\Pi$.
\end{defin}

Intuitively, an answer set represents one possible meaning of knowledge encoded by logic 
program $\Pi$ and a set of classical literals $S$. In a typical case like ours, one answer set
{\em corresponds to one valid solution of a problem} encoded by our logic program. If the ASP solver returns
more than one answer set, there is more than one solution. If it fails to return any answer sets, no
solution to this problem exists.

\section{Encoding the Problem in ASP}\label{section:encoding}

For our implementation, we were using a modern ASP solver called {\em clingo}\footnote{\tt http://potassco.sourceforge.net/}.
It supports an extended modelling language, described in \cite{Gebser-08} and \cite{Gebser-09}. 
In addition to basic ASP constructs ({\em rules, facts, constraints}), it has a support for
{\em generator rules}, {\em optimization statements} and has a set of built-in {\em arithmetic functions} and {\em aggregates},
making the problem definition more convenient for us.

Our bot needed to prepare a logic program describing the current problem instance (finding a wall-in placement with a given
set of buildings at a certain chokepoint), pass it to the solver, read the results from standard output and 
parse them to obtain resulting tile positions. The solver accepts this logic program either from standard input, or it can 
be read from a text file (command to execute is then: ``{\tt clingo findWallPlacement.txt}" \footnote{In our implementation,
we call the ``{\tt clingo --asp09 findWallPlacement.txt}" command. The {\tt --asp09} flag instructs {\em clingo} to
display results in the format of ASP Competition'09, which is easier to parse.}). Contents of the {\tt findWallPlacement.txt}
file will be described in the rest of this section.
Typical output returned by the solver is depicted in figure \ref{fig:output}.

\vskip 5mm
\begin{figure}[h!]
{
\ttfamily\scriptsize
\begin{verbatim}
Answer: 1
place(zealot,126,23) place(pylon,122,22) place(gateway,118,23) place(forge,124,23)
Optimization: 42 0
OPTIMUM FOUND
\end{verbatim}
}
\caption{Output of the ASP solver corresponding to the wall from figure \ref{fig:wall-toss1}.}
  \label{fig:output}
\end{figure}

One call of the ASP solver with our problem encoding takes less than 200 miliseconds on a single-CPU 
virtual machine running Windows XP with 1GB of available RAM. 
This allows us to attempt to compute the wall placement with fewer buildings at first, and then keep adding 
more of them, until the wall-in is possible\footnote{The wall-in is possible, if the solver returns some 
non-zero number of answer sets. If it fails to return any, it means that the given chokepoint cannot 
be blocked by a given set of buildings.}. In the case from figure \ref{fig:wall-toss1}, we first tried to compute the wall with only a 
Gateway, Pylon and a Zealot. Since this was not possible, we tried adding a Forge to our logic program 
and called the solver again (this time with success).

\subsection{Buildings to Use}\label{subsection:buildings}
The first part of our logic program in {\tt findWallPlacement.txt} file contains an encoding of the 
buildings we want to use. First of all, we use BWAPI to generate the following set of facts defining the building types, 
their sizes and gaps around them (notice the fake ``zealots" building type).

{\scriptsize
\color{MidnightBlue}
\begin{verbatim}
% Specify building types, their sizes and gaps.
buildingType(forgeType).
buildingType(gatewayType).
buildingType(pylonType).
buildingType(zealotsType).

width(gatewayType,4). height(gatewayType,3).	
width(forgeType,3).   height(forgeType,2).		
width(pylonType,2).   height(pylonType,2).		
width(zealotsType,1). height(zealotsType,1).	

leftGap(gatewayType,16). rightGap(gatewayType,15). topGap(gatewayType,16). bottomGap(gatewayType,7).
leftGap(forgeType,12).   rightGap(forgeType,11).   topGap(forgeType,8).    bottomGap(forgeType,11).
leftGap(pylonType,16).   rightGap(pylonType,15).   topGap(pylonType,20).   bottomGap(pylonType,11).
leftGap(zealotsType,0).  rightGap(zealotsType,0).  topGap(zealotsType,0).  bottomGap(zealotsType,0).
\end{verbatim}
}

After that, we need to specify the building instances which we want to have in our wall and assign them to their 
corresponding types. This can be done with the following collection of ASP facts:

{\scriptsize
\color{MidnightBlue}
\begin{verbatim}
% Specify what building instances to build.
building(pylon1).
type(pylon1,pylonType).
building(forge1).
type(forge1,forgeType).
building(gateway1).
type(gateway1,gatewayType).
building(zealots).
type(zealots1,zealotsType).
\end{verbatim}
}

Following constraint simply states that {\em two different buildings must not occupy
the same tile position} (must not overlap). Note that the ``$\leftarrow$" symbol from the rule definition above
is written as ``{\tt :-}" in our text file. Also, the convention in ASP is that 
constants ({\em forge1, pylonType}) start with lowercase letters or numbers, while the 
variables are uppercase ({\em B2, X, Y}).

\vskip 4mm
{\scriptsize
\color{MidnightBlue}
\begin{verbatim}
% Constraint: Two buildings cannot occupy the same tile.
:- occupiedBy(B1,X,Y), occupiedBy(B2,X,Y), B1!=B2.
\end{verbatim}
}

However, we still need to specify which tile positions are occupied by which buildings. 
This is taken care of by the following rule. Here, we use previously established 
description of the {\em type, width} and {\em height} of our buildings. The ``$place$"
literal is especially important and will be explained in subsection \ref{section:generate-and-test}.
Intuitively, this rule says that {\em if we place building $B$ on tile $(X_1,Y_1)$
and $X_1 \leq X_2 < X_1+$ width\_of\_$B$, and at the same time $Y_1 \leq Y_2 < Y_1+$ height\_of\_$B$, then 
also the $(X_2,Y_2)$ tile is occupied by $B$.}

\vskip 5mm
{\scriptsize
\color{MidnightBlue}
\begin{verbatim}
% Tiles occupied by the buildings.
occupiedBy(B,X2,Y2) :- place(B,X1,Y1), 
                       type(B,BT), width(BT,Z), height(BT,Q),
                       X2 >= X1, X2 < X1+Z, Y2 >= Y1, Y2 < Y1+Q, 
                       walkableTile(X2,Y2).
\end{verbatim}
}

Following four rules simply compute the horizontal and vertical gaps between every pair of
adjacent tiles occupied by different buildings. The $verticalGap$ and $horizontalGap$ literals
will be used during the optimization.

\vskip 5mm
{\scriptsize
\color{MidnightBlue}
\begin{verbatim}
% Gaps between every two adjacent tiles, that are occupied by buildings.
verticalGap(X1,Y1,X2,Y2,G) :-
        occupiedBy(B1,X1,Y1), occupiedBy(B2,X2,Y2), 
        B1 != B2, X1=X2, Y1=Y2-1, G=S1+S2, 					
        type(B1,T1), type(B2,T2), bottomGap(T1,S1), topGap(T2,S2).
    
verticalGap(X1,Y1,X2,Y2,G) :-
        occupiedBy(B1,X1,Y1), occupiedBy(B2,X2,Y2), 
        B1 != B2, X1=X2, Y1=Y2+1, G=S1+S2,
        type(B1,T1), type(B2,T2), bottomGap(T2,S2), topGap(T1,S1).
    
horizontalGap(X1,Y1,X2,Y2,G) :-
        occupiedBy(B1,X1,Y1), occupiedBy(B2,X2,Y2), 
        B1 != B2, X1=X2-1, Y1=Y2, G=S1+S2,
        type(B1,T1), type(B2,T2), rightGap(T1,S1), leftGap(T2,S2).
    
horizontalGap(X1,Y1,X2,Y2,G) :-
        occupiedBy(B1,X1,Y1), occupiedBy(B2,X2,Y2), 
        B1 != B2, X1=X2+1, Y1=Y2, G=S1+S2,
        type(B1,T1), type(B2,T2), rightGap(T2,S2), leftGap(T1,S1).
\end{verbatim}
}

\subsection{Terrain Encoding}
Now we need to express what the terrain around the chokepoint looks like. 
Specifically, we need to have a set of facts describing which tile positions are walkable
and where there is enough space to build individual building types. Information about this
is easily accessible via BWAPI. 

{\scriptsize
\color{MidnightBlue}
\begin{verbatim}
walkableTile(87,11).
buildable(pylonType,86,8).
buildable(gatewayType,90,8).
buildable(pylonType,85,19).
buildable(pylonType,88,13).
walkableTile(94,21).
buildable(zealotsType,93,10).
buildable(pylonType,88,19).
walkableTile(94,9).
buildable(gatewayType,85,11).
walkableTile(88,9).
buildable(pylonType,89,11).
walkableTile(94,18).
walkableTile(89,21).

...etc.
\end{verbatim}
}

\begin{figure}[h!]
  \begin{center}
    \includegraphics[width=0.5\textwidth]{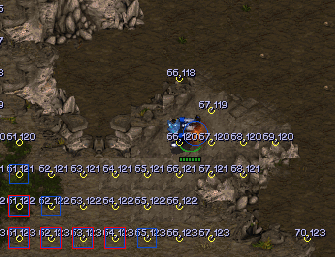}
  \end{center}
  \caption{Terrain around the chokepoint. Blue squares denote the tiles where a
  Forge can be built and red squares the tiles where we can build a Gateway.}
  \label{fig:terrain}
\end{figure}

To keep the computation times low, we only take into account the tiles from a certain 
area around the chokepoint. In our implementation, we considered a 16-tile radius 
around the chokepoint's center.

\newpage
In addition to that, we also need to have a specification of two important tile positions - one inside
our base region and the other one outside of it. These will be used when determining if the wall blocks the passage or not.
If there is no walkable path between these two positions, the wall is tight.

\vskip 5mm
{\scriptsize
\color{MidnightBlue}
\begin{verbatim}
insideBase(94,10).
outsideBase(84,22).
\end{verbatim}
}

For the {\em insideBase} position, we simply select the tile from within our considered radius that is 
closest to our base. Similarly, the {\em outsideBase} is a tile position that is closest to the center of the outer
region.

\subsection{Reachability}

The following constraint says that units must not be able to reach the 
base region (specifically position $X_{2},Y_{2}$) from the outside region (position $X_{1},Y_{1}$).

\vskip 5mm
{\scriptsize
\color{MidnightBlue}
\begin{verbatim}
% Constraint: Inside of the base must not be reachable.
:- insideBase(X2,Y2), outsideBase(X1,Y1), canReach(X2,Y2).
\end{verbatim}
}

However, we still need to declare which tiles are reachable. The following set of rules basically takes care 
of the pathfinding problem. First of them says that a walkable position is blocked, if it is occupied by a building.
Second rule states that the {\em outsideBase} position is reachable from itself (obviously). 
A collection of eight rules after that recursively identifies a position as {\em reachable} if it is {\em not blocked}, and 
it {\em has a reachable neighbor} in any of the eight directions. 

\vskip 5mm
{\scriptsize
\color{MidnightBlue}
\begin{verbatim}
% Reachability between tiles.
blocked(X,Y) :- occupiedBy(B,X,Y), building(B), walkableTile(X,Y).
canReach(X,Y) :- outsideBase(X,Y).

canReach(X2,Y) :- 
      canReach(X1,Y), X1=X2+1, walkableTile(X1,Y), walkableTile(X2,Y), 
      not blocked(X1,Y), not blocked(X2,Y).
canReach(X2,Y) :- 
      canReach(X1,Y), X1=X2-1, walkableTile(X1,Y), walkableTile(X2,Y), 
      not blocked(X1,Y), not blocked(X2,Y).
canReach(X,Y2) :- 
      canReach(X,Y1), Y1=Y2+1, walkableTile(X,Y1), walkableTile(X,Y2), 
      not blocked(X,Y1), not blocked(X,Y2).
canReach(X,Y2) :- 
      canReach(X,Y1), Y1=Y2-1, walkableTile(X,Y1), walkableTile(X,Y2), 
      not blocked(X,Y1), not blocked(X,Y2).
canReach(X2,Y2) :- 
      canReach(X1,Y1), X1=X2+1, Y1=Y2+1, walkableTile(X1,Y1), walkableTile(X2,Y2), 
      not blocked(X1,Y1), not blocked(X2,Y2).
canReach(X2,Y2) :- 
      canReach(X1,Y1), X1=X2-1, Y1=Y2+1, walkableTile(X1,Y1), walkableTile(X2,Y2), 
      not blocked(X1,Y1), not blocked(X2,Y2).
canReach(X2,Y2) :- 
      canReach(X1,Y1), X1=X2+1, Y1=Y2-1, walkableTile(X1,Y1), walkableTile(X2,Y2), 
      not blocked(X1,Y1), not blocked(X2,Y2).
canReach(X2,Y2) :- 
      canReach(X1,Y1), X1=X2-1, Y1=Y2-1, walkableTile(X1,Y1), walkableTile(X2,Y2), 
      not blocked(X1,Y1), not blocked(X2,Y2).
\end{verbatim}
}

There is, however, one more source of reachability in StarCraft - gaps at the sides of the 
buildings. Because of them, the blocked locations may sometimes be reachable too.
To make our problem representation more precise, we should take them into account as well.
In section \ref{subsection:buildings}, we computed the horizontal and vertical gaps between
the pairs of tile positions, which will be used now. 

First of all, we specify a pixel dimensions of an enemy unit type that we want to lock outside of our base 
(in this case, it is the smallest unit in the game: $16 \times 16$px Zergling).
Then we use the following eight rules to declare more positions reachable (from corresponding 
directions) if the gap is wide enough, even if these positions are blocked.

\vskip 5mm
{\scriptsize
\color{MidnightBlue}
\begin{verbatim}
% Using gaps to reach (walk on) blocked locations.
enemyUnitX(16). enemyUnitY(16).
canReach(X1,Y1) :- horizontalGap(X1,Y1,X2,Y1,G), G >= S, X2=X1+1, canReach(X1,Y3), Y3=Y1+1, enemyUnitX(S).
canReach(X1,Y1) :- horizontalGap(X1,Y1,X2,Y1,G), G >= S, X2=X1-1, canReach(X1,Y3), Y3=Y1+1, enemyUnitX(S).
canReach(X1,Y1) :- horizontalGap(X1,Y1,X2,Y1,G), G >= S, X2=X1+1, canReach(X1,Y3), Y3=Y1-1, enemyUnitX(S).
canReach(X1,Y1) :- horizontalGap(X1,Y1,X2,Y1,G), G >= S, X2=X1-1, canReach(X1,Y3), Y3=Y1-1, enemyUnitX(S).
canReach(X1,Y1) :- verticalGap(X1,Y1,X1,Y2,G), G >= S, Y2=Y1+1, canReach(X3,Y1), X3=X1-1, enemyUnitY(S).
canReach(X1,Y1) :- verticalGap(X1,Y1,X1,Y2,G), G >= S, Y2=Y1-1, canReach(X3,Y1), X3=X1-1, enemyUnitY(S).
canReach(X1,Y1) :- verticalGap(X1,Y1,X1,Y2,G), G >= S, Y2=Y1+1, canReach(X3,Y1), X3=X1+1, enemyUnitY(S).
canReach(X1,Y1) :- verticalGap(X1,Y1,X1,Y2,G), G >= S, Y2=Y1-1, canReach(X3,Y1), X3=X1+1, enemyUnitY(S).
\end{verbatim}
}

Now we should have our terrain, building instances, reachability, and all the constraints ready, so we 
can finally start generating solutions.

\subsection{Generate and Test Method}\label{section:generate-and-test}

Our logic program uses the ``{\em generate-and-test}" organization that is very often employed in ASP-based problem solving.
The idea is simple: we use so-called {\em generator rules} (or choice rules if generators are not supported by our 
solver) to describe a large set of ``{\em potential solutions}" to our problem. In our case, any building placement
is a potential solution.
Each of them is then tested - we check if all the constraints are satisfied in it. 
If there are some constraints, that are not satisfied, the solution is thrown away. Otherwise, it is {\em returned as an 
answer set}.

The generator rules for our wall-in placement problem are depicted below (there is one rule for every 
building instance\footnote{Note that it is also possible to write a single generator rule for all the building instances. 
We have chosen the encoding with multiple generator rules for this text, because we feel it is more intuitive.}). 
For example, the first rule says that ``for every $X$ and $Y$, such that $forgeType$ is
buildable on $(X,Y)$, we want to generate and check potential solutions with exactly one literal $place(forge1,X,Y)$" (exactly one,
because we want every building instance to be placed, and no building can be placed on more than one position 
at the same time).

\vskip 5mm
{\scriptsize
\color{MidnightBlue}
\begin{verbatim}
% Generate all the potential placements.
1[place(forge1,X,Y) : buildable(forgeType,X,Y)]1.
1[place(pylon1,X,Y) : buildable(pylonType,X,Y)]1.
1[place(gateway1,X,Y) : buildable(gatewayType,X,Y)]1.
1[place(zealots,X,Y) : buildable(zealotsType,X,Y)]1.
\end{verbatim}
}

On the output, the ASP solver returns the answer sets corresponding to solutions with exactly one $place$
literal for every building instance, where all the constraints are satisfied.

\subsection{Optimization}
As we mentioned before, there might be more valid solutions to our problem, some of which are better than
others. We consider the wall-in better, if there are less wide gaps in it. 
The ASP solver can be instructed to find the best among all the valid solutions 
by including the following two optimization statements in our logic program.

\vskip 5mm
{\scriptsize
\color{MidnightBlue}
\begin{verbatim}
% Optimization statements.
#minimize [verticalGap(X1,Y1,X2,Y2,G) = G ].
#minimize [horizontalGap(X1,Y1,X2,Y2,G) = G ].
\end{verbatim}
}

Note that we can omit the optimization statements if we do not necessarily need the best possible solution. 
By doing this, we can save some computation time, since computing any answer set is faster than 
computing the best one.

\vskip 10mm
Logic programs, like the one described in this section, can be easily generated
by any bot programmed in BWAPI and used to find a wall-in placement for most of the chokepoints
in typical StarCraft tournament maps. The computation time is bearable, considering that the bots 
do not need to solve this problem very frequently (usually only once per game).

\section{Summary}\label{section:summary}

We have shown how a declarative programming approach can be used to generate wall-in 
placements in StarCraft (and possibly other RTS games) quite effortlessly. 

This text, written as a guide for bot programmers, addresses all the relevant subproblems, like 
terrain and building encoding, pathfinding/reachability and optimization, and presents a declarative 
solution to each of them. 
The ASP encoding, presented here and employed in our implementation, can be directly adopted by 
other bot programmers. However, we encourage readers to experiment with different declarative paradigms
and technologies.


\bibliographystyle{apalike}


\end{document}